\documentclass[letterpaper, 10 pt, conference]{ieeeconf}  % Comment this line out if you need a4paper

\IEEEoverridecommandlockouts                              % This command is only needed if 
\usepackage[english]{babel}

\usepackage{hyperref}
\usepackage[]{graphicx}
\usepackage[caption=false]{subfig}
\usepackage{bm}
\usepackage{eucal}
\usepackage{dsfont}
\usepackage{mathtools}
\usepackage[export]{adjustbox}
\usepackage[OT1]{fontenc}
\usepackage{physics}
\usepackage{algorithm}
\usepackage{algpseudocode}
\usepackage{nicefrac}
\usepackage{optidef}
\usepackage{siunitx}
\usepackage{todonotes}
\usepackage{accents}
\usepackage{booktabs}
\usepackage[capitalise]{cleveref}
\usepackage{lipsum}
\usepackage{tikz}
\usepackage{multirow}

\usepackage{csquotes}
\usepackage[style=ieee, citestyle=numeric-comp, backend=biber, url=false]{biblatex}

\usetikzlibrary{graphs}

\crefformat{equation}{(#2#1#3)}
\crefrangeformat{equation}{(#3#1#4) to~(#5#2#6)}
\crefmultiformat{equation}{(#2#1#3)}%
{ and~(#2#1#3)}{, (#2#1#3)}{ and~(#2#1#3)}

\newcommand{\set}[1]{\mathbf{#1}}

\newcommand{\dist}[1]{\mathcal{#1}}
\newcommand{\func}[1]{\mathds{#1}}
\newcommand{\transp}[1]{{#1}^{\mathrm{T}}}

\newcommand{\inv}[1]{{#1}^{-1}}
\newcommand{\define}{\coloneqq}

\usepackage{dblfloatfix}

\title{\LARGE \bf
Operator Splitting Covariance Steering
\\
for Safe Stochastic Nonlinear Control
}

\author{Akash Ratheesh,\textsuperscript{\textdagger} Vincent Pacelli,\textsuperscript{\textdagger} Augustinos D. Saravanos and Evangelos A. Theodorou% <-this % stops a space
\thanks{This work is supported by the National Aeronautics and Space Administration under ULI Grant 80NSSC22M0070. Augustinos Saravanos acknowledges support by the A. Onassis Foundation Scholarship.}% <-this % stops a space
\thanks{Akash Ratheesh, Vincent Pacelli and Evangelos Theodorou are with the School of Aerospace Engineering, Georgia Institute of Technology, Atlanta, GA, USA.
{\tt\footnotesize \{akashratheesh, vpacelli, evangelos.theodorou\}@gatech.edu}}%
\thanks{Augustinos Saravanos is with the School of Electrical and Computer Engineering, Georgia Institute of Technology, Atlanta, GA, USA.
{\tt\footnotesize asaravanos@gatech.edu}}%
\thanks{\textsuperscript{\textdagger} Equal Contribution.}%
}

\begin{document}

\maketitle
\thispagestyle{empty}
\pagestyle{empty}

% TODO List
% 1. Figures
% 2. Merge tables?
% 3. Finalize experiment discussion.
% 4. Write conclusion
% 5. Cut down.

%%%%%%%%%%%%%%%%%%%%%%%%%%%%%%%%%%%%%%%%%%%%%%%%%%%%%%%%%%%%%%%%%%%%%%%%%%%%%%%%
\begin{abstract}
This paper presents a novel algorithm for solving distribution steering problems featuring nonlinear dynamics and chance constraints. Covariance steering (CS) is an emerging methodology in stochastic optimal control that poses constraints on the first two moments of the state distribution --- thereby being more tractable than full distributional control. Nevertheless, a significant limitation of current approaches for solving nonlinear CS problems, such as sequential convex programming (SCP), is that they often generate infeasible or poor results due to the large number of constraints. In this paper, we address these challenges, by proposing an operator splitting CS approach that temporarily decouples the full problem into subproblems that can be solved in parallel. This relaxation does not require intermediate iterates to satisfy all constraints simultaneously prior to convergence, which enhances exploration and improves feasibility in such non-convex settings. Simulation results across a variety of robotics applications verify the ability of the proposed method to find better solutions even under stricter safety constraints than standard SCP. Finally, the applicability of our framework on real systems is also confirmed through hardware demonstrations (Video: \url{https://youtu.be/UCyYcDITO2Q}).

\end{abstract}

\section{Introduction}

% Still working on this.

In modern control applications, autonomous systems must satisfy a number of specified design and performance criteria with a high degree of confidence to ensure their reliable performance. For this reason, controlling distributions of system trajectories via stochastic optimal control (SOC) is a promising field for meeting such performance specifications \cite{Renardy06, Bakolas18, Okamoto19, Balci24, Wu23a}. Nevertheless, fully controlling the density of the state distribution is an infinite-dimensional problem which suffers from the curse of dimensionality \cite{Renardy06}. As such, developing methods that can effectively control the distributions of nontrivial systems while remaining scalable and computationally efficient remains a significant challenge for safe autonomy.

%In applications like include self-driving cars and mobile robots \cite{teng2023motion}, aircraft landing \cite{gautam2014survey}, manipulators interacting with their environment \cite{alandoli2020critical}, and power systems \cite{Khalil25} it is important to synthesize controllers with safety guarantees under uncertainty.

One emerging paradigm for synthesizing such controllers is covariance steering (CS) --- a SOC methodology that enforces constraints on the mean and covariance of the state \cite{Hotz1987, Chen15, Bakolas18, Balci24, Okamoto19}. Successful applications of CS can be found in trajectory optimization \cite{Balci22, Yin22}, path planning \cite{Okamoto19, Knaup23a}, manipulation \cite{Lee22}, power systems \cite{Khalil25}, and large-scale multi-agent control \cite{Saravanos21, Saravanos23}. Since CS only concerns the first two moments of the system state, it is computationally tractable in comparison to density control problems. Although CS originally focused on unconstrained linear systems, CS problems (CSPs) under nonlinear dynamics and chance constraints can be solved via iterative local approximations using algorithms like sequential convex programming (SCP) \cite{Ridderhof19, Yu2021, Saravanos22}.
% , much like differential dynamic programming (DDP) \cite{murray84, Tassa07}. 
% While the resulting solutions only provide local information about the first two moments of the system, this is sufficient information to estimate trajectory statistics via concentration inequalities \cite{Yu2021, Hoeffding94, Hanson71, Rudelson13}.

\begin{figure}[t]
    \centering
    \includegraphics[width=\columnwidth]{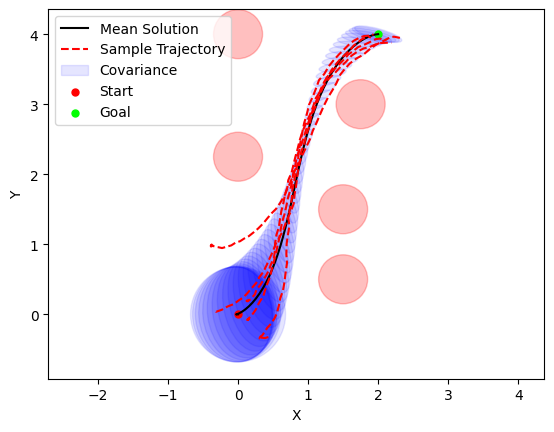}
    \vspace{-0.8cm}
    \caption{The proposed algorithm steering a unicycle robot from an initial to a target distribution under probabilistic obstacle avoidance constraints. 
    % (described in \cref{sec: sim results - unicycle}) 
    % Note that the controller is capable of successfully steering samples outside of the 99\% confidence region.
    }
    \label{fig:unicycle}
\vspace{-0.5cm}
\end{figure}

While CS is effective at designing controllers that explicitly model uncertainty, solving chance-constrained, nonlinear CSPs remains difficult. The approximations required to formulate tractable constraints at each step of iterative algorithms tend to be very conservative. As a result, heavily constrained problems (e.g., motion planning with many obstacles) challenge existing SCP methods that may erroneously eliminate feasible minima, degrading performance or causing the solver to incorrectly report problems as infeasible. Such issues further compound over iterations of SCP algorithms as they require repeated conservative approximations.

This paper presents a novel CS algorithm for problems featuring nonlinear dynamics and state chance constraints, such as those imposed by workspace obstacles. The proposed method is derived based on an operator splitting formulation and solved using the alternating direction method of multipliers (ADMM) \cite{Boyd11}. Notably, the algorithm is tolerant to infeasible iterates before converging to a solution. This property allows the proposed algorithm to find solutions that feature superior costs under tighter constraints than traditional solvers on the same CSPs. These improvements are particularly important for applications that require trajectories to satisfy many state space constraints along with other performance requirements. The efficacy of the proposed algorithm is shown via extensive simulation of both linear and nonlinear motion planning problems with nonconvex state spaces as well as hardware experiments.

\textbf{Summary of Contributions.} Our first main contribution is the formulation of an operator splitting algorithm for covariance steering problems involving nonlinear dynamics and state chance constraints. The method relies on an iterative approximation scheme for forming a sequence of convex optimization problems that are then decoupled into smaller subproblems which are solved in parallel with ADMM. This relaxation allows for intermediate iterates to be temporarily infeasible, thus enhancing the exploration capabilities of the algorithm in contrast with existing methods. Subsequently, we demonstrate in practice 
the ability of this approach to find solutions superior to those produced by traditional solvers in extensive simulations of various linear and nonlinear motion planning problems. 
% This improvement is due to the fact that, since the ADMM is tolerate to infeasible initializations and iterations, it is capable of finding superior solutions under tighter constraints than traditional optimizers. 
Finally, hardware experiments highlight the effectiveness of the proposed CS algorithm in controlling real systems.

\section{Related Work}
% \subsection{Covariance Steering}

Controlling distributions of systems is a promising field for achieving safety requirements that are common in control and robotics \cite{Renardy06, Bakolas18, Okamoto19, Balci24, Wu23a, Wu23b}. In this direction, covariance steering has emerged as a tractable methodology for explicitly imposing constraints on the mean and covariance of the system state under stochastic uncertainty \cite{Hotz1987, Chen15, Bakolas18, Balci24, Okamoto19}. This approach significantly differs from standard SOC approaches, e.g., linear-quadratic-Gaussian (LQG) control \cite{Stengel94}, which indirectly control the uncertainty through the cost function. 
% where the requirement to reduce system uncertainty only appears as higher moments of the control cost, thereby implicitly controlling uncertainty unlike in CS problem. %Therefore, in traditional SOC problems, uncertainty can only be controlled implicitly through shaping the cost function, and not explicitly as in CS problems.

CS methods were initially formulated and studied under an infinite horizon setting during the 1980s \cite{Collins85, Hotz1987}. At that time, their finite-horizon counterparts were most likely ignored due to the necessity of solving associated semidefinite programs (SDP), which was computationally expensive. Advances in computational resources and optimization techniques, however, have led to finite-horizon CS methods receiving significantly renewed interest \cite{Yu2021, Knaup23b, Ridderhof19, Okamoto19, Pilipovsky21, Balci24, Rapakoulias23, Rapakoulias24, Balci22, Chen21, Chen20, Liu22, Yin22, Saravanos21, Saravanos22, Saravanos23} as a practical alternative to controlling the full density of the state distribution \cite{Chen21, Moyalan21}. Note that the latter approaches produce infinite-dimensional optimization problems which require the approximate solution of PDEs, thus reducing their scalability. In contrast, CS methods have been shown to provide solutions in reasonable computational time even for large-scale systems with hundreds to millions of states when combined with modern optimization techniques \cite{Saravanos21, Saravanos23}. Robotics applications of CS include path planning \cite{Okamoto19, Knaup23a}, manipulation \cite{Lee22}, soft landing \cite{Ridderhof18} and multi-agent control \cite{Saravanos21, Saravanos22} to name only a few. 
 
% Moreover, although CS methods only uses information local to the mean trajectory, they provide estimates of the cost and other statistics via concentration inequalities \cite{Yu2021}, like the Hoeffding \cite{Hoeffding94} or Hanson-Write bounds \cite{Hanson71, Rudelson13}. 

Many recent methods have expanded CS problems to include nonlinear dynamics \cite{Ridderhof19, Yu2021, Saravanos22} and chance constraints \cite{Balci24, Pilipovsky21, Rapakoulias23, Okamoto19} through iterative approximation schemes. Such modifications are particularly relevant to safety critical systems, where the system entering certain states can lead to permanent damage. Nevertheless, the inclusion of nonlinear dynamics and chance constraints complicate CS problems since they require conservative approximations to produce a convex subproblem during each iteration of the algorithm. In fact, these approximations might limit the ability of such approaches to find high quality --- or even feasible --- solutions as the amount of constraints increases. 
% This issue is even compounded in iterative methods when locally approximating such constraints. 
% After updating the trajectory about which approximation occurs, the new trajectory can become infeasible for the new CSP, causing many convex optimization algorithms to fail. 
Current literature avoids this problem by only considering linear state constraints that produce a convex polytope of feasible states \cite{Okamoto19, Knaup23a, Pilipovsky21}. Nonconvex state spaces are then managed by decomposing them into convex polytopes, then employing more expensive mixed-integer programming methods to find trajectories across polytope boundaries \cite{Okamoto19}. In contrast, this paper presents an algorithm for effectively controlling systems with nonlinear dynamics and nonconvex state constraints. Due to our method's ability to tolerate infeasibility in iterates prior to convergence, it is able to handle nonconvex state constraints directly without decomposing the state space into convex regions.

% \subsection{Operator Splitting Schemes in Robotics}

In contrast to other methods for nonlinear CS problems, our algorithm relies on an operator splitting method. Operator splitting is a numerical technique where a function can be decomposed into two or more subproblems and a specialized method is used to tackle each subproblem in an alternating manner. While originally used for solving differential equations, it is also applicable to optimization problems. Operator splitting approaches using the ADMM are becoming increasingly popular methods to solve optimal control problems due to their ability to decouple complex problems into simpler subproblems that can be solved in parallel \cite{Sindhwani17, Saravanos23b}. Several successful applications can be found in trajectory optimization \cite{Sindhwani17, Wang21}, model predictive control \cite{Rey20},  multi-agent control \cite{Saravanos23b, Pereira22, Chen2021b, Ferranti22, Schwager22}, manipulation \cite{Aydinoglu22}, legged locomotion \cite{Zhou20, Budhiraja19}, etc. 

\section{Nonlinear Covariance Steering \\ Problem Formulation}
\subsection{Notation}

Let $X$ be a random variable (r.v.). The expected value and covariance of $X$ are denoted $\func{E}[X]$ and $\func{C}[X]$ respectively. If a r.v. $x \sim \dist{N} (\mu, \Sigma)$, then it is subject to a Gaussian distribution with mean $\mu$ and covariance $\Sigma$.  The set $\set{S}^n_+$ ($\set{S}^n_{++}$) denotes the set of symmetric, positive semidefinite (definite) matrices with $n$ rows and columns. For two matrices $A, B$, the notation $A \succeq B$ indicates that $A - B \in \set{S}^n_+$. For a function $f(x)$, its gradient at a point $\bar{x} \in \set{R}^n$ is $\partial_x f(\bar{x})$. Similarly, if $f(x)$ is real-valued, its Hessian at $\bar{x}$ is denoted $\partial_x^2 f(\bar{x})$. All functions are assumed to be sufficiently smooth for the required derivatives to exist. Finally, we define the proximal operator:
\begin{align}{\operatorname{prox}}_{\rho, f}(z) =  {\operatorname{argmin}_x} f(x) + \frac{\rho}{2} \| x - z\|_2^2.
\end{align}

\subsection{Problem Formulation}

Our method applies to discrete-time nonlinear dynamical systems with Gaussian process noise, i.e.,
% \footnote{Time-varying dynamics and costs are also allowed. We do not consider this case explicitly to keep notation compact.}
\begin{align}
    X_{t + 1} = f(X_t, u_t) + D(X_t) W_t, \label{eq:dynamics}
\end{align}
where $X_t$ is the r.v. representing the system state, $u_t$ is the control to be determined, $f(x, u)$ are the deterministic dynamics, $D(x)$ is a state-dependent diffusion coefficient, and $W_t \sim \dist{N}(0, I)$ is a standard Gaussian. 
% The random variable $W_t$ is noise distributed according to a standard Gaussian. 
The initial and terminal conditions are given by:
\begin{align}
\func{E}[X_0] & = \mu_{\mathrm{ic}}, & \func{C}[X_0] & = \Sigma_{\mathrm{ic}}, 
\nonumber
\\
\func{E}[X_{t_f}] & = \mu_{\mathrm{tc}}, & \func{C}[X_{t_f}] & = \Sigma_{\mathrm{tc}}.
\label{eq:boundary conditions}
\end{align}

The state is subject to $N$ nonlinear constraints, (e.g., workspace obstacles) defined by functions $h_i(x)$. Each such function $h_i(x)$ defines a set $\set{H}_i \define \{x\ |\ h_i(x) \leq 0\}$. Since the support of $X_t$ is not compact, it is impossible to satisfy these constraints almost surely. This requirement is relaxed to only hold with high probability via the chance constraint:
\begin{align}
    \func{P}\{X_t \in \set{H}\} \leq \delta, && \set{H} \define \bigcup_{i = 1}^N \set{H}_i. \label{eq:chance constraint}
\end{align}

Finally, the controls are selected to minimize in expectation a function $c(x)$ plus a quadratic control defined by $R \in \set{S}^n_{++}$ tallied at each time step $t$. Therefore, the complete covariance steering problem is formulated as:
\footnote{
% The algorithm described in the next section applies to a broader class of problems as well. 
The terminal covariance constraint can also be relaxed to an inequality constraint, i.e., $\func{C}[X_{t_f}] \preceq \Sigma_{\mathrm{tc}}$, which indicates that the terminal covariance of the system simply needs to be ``smaller'' than $\Sigma_{\mathrm{tc}}$. 
% This relaxation is often sufficient in robotics applications, since the goal is to drive the robot to be within a region with a high probability, not to exactly match a distribution. This relaxation can also improve feasibility. Other possible modifications include applying constraints to the mean or covariance at other time steps, removing the mean terminal constraint if it is not needed, and a general differentiable cost $c(x, u)$. Only the formulation in \cref{eq:nonlinear cov steer prob} is treated explicitly for brevity.
}
\begin{mini}|s|
    {u_{0:t_f - 1}}{\sum_{t = 0}^{t_f - 1} \func{E}\qty[c(X_t) + \frac{1}{2}\transp{u}_t R u_t],}
    {\label{eq:nonlinear cov steer prob}}{}
    \addConstraint{\cref{eq:dynamics}, \cref{eq:boundary conditions}. \cref{eq:chance constraint}}{}{}
\end{mini}

\section{Chance Constrained Covariance Steering \\ via Operator Splitting}
This section formulates a new algorithm for solving chance-constrained CSPs. It is organized as follows. First, the controller parameterization is described, which has a significant impact on the CSP. Next, a local convex approximation to \cref{eq:nonlinear cov steer prob} is formulated and the operator splitting scheme is described. Finally, the forward pass used to update the local approximation and complete algorithm are presented.

\subsection{Controller Parameterization}
\label{ssec:controller parameterization}
An important aspect of solving \cref{eq:nonlinear cov steer prob} is the parameterization of the $u_t$. The choice of parameterization impacts the convex approximation of the chance constraint \cite{Balci24}, the number of decision variables, and how the feedback is performed. As our method relies on iterative linear-quadratic approximations, we select a linear state feedback of the form:
\begin{align}
    u_t(x) = v_t + K_t (x - \mu_t). \label{eq:control law}
\end{align}
The benefits of this formulation are that the number of decision variables is comparatively small, the feedback is determined directly by the deviation from the mean trajectory to help correct for nonlinearities during the forward pass, and projections onto the chance constraint set can be computed in parallel for each time step. The primary disadvantage is that the common approximation of the chance constraint as a second-order cone constraint is not possible, leading to a more conservative approximation.

Other options present in the literature differ in how they use feedback to control the covariance \cite{Balci24}. The choices include applying linear feedback to histories of states, histories of the disturbances \cite{Balci24}, or to the state of a zero-mean auxiliary stochastic process representing the impact of the noise on the system \cite{Ridderhof19, Pilipovsky21}. For most of these formulations, the chance constraint convex approximation is tighter, but lack the benefits of the selected parameterization.

\subsection{Local Convex Approximation of the CS Problem}
This subsection performs a local approximation of \cref{eq:nonlinear cov steer prob} using a nominal trajectory $\tau = (\bar{x}_{0:t_f}, \bar{u}_{0:t_f}, \bar{\Sigma}_{0:t_f})$ satisfying $\bar{x}_{t + 1} = f(\bar{x}_t, \bar{u}_t)$. In \cref{ssec:trajectory update}, how this trajectory is updated from an initial guess is discussed.

\textbf{Dynamics.} In order to produce local solutions to the covariance steering problem \cref{eq:nonlinear cov steer prob}, a first-order Taylor expansion is performed, yielding affine dynamics,
\begin{align}
X_{t + 1} = A_t X_t + B_t u_t + D_t w_t + d_t,\ w_t \sim \dist{N}(0, I),
\end{align}
where $A_t \define \partial_x f(\bar{x}_t, \bar{u}_t), B_t \define \partial_u f(\bar{x}_t, \bar{u}_t), D_t \define D(\bar{x}_t),$ and $d_t \define f(\bar{x}_t, \bar{u}_t) - A_t \bar{x}_t - B_t \bar{u}_t$. Under the feedback law \cref{eq:control law}, the state distribution remains Gaussian and the mean and covariance evolve as:
\begin{subequations}
\begin{align}
\mu_{t + 1} &= A_t \mu_t + B_t v_t + d_t, \label{eq:mean dynamics}\\
\Sigma_{t + 1} &= (A_t + B_t K_t) \Sigma_t \transp{(A_t + B_t K_t)} + D_t \transp{D}_t,\label{eq:cov dynamics}
\end{align}
\end{subequations}
Define the dynamically feasible sets of $\mu_{0:t_f}$ and $\Sigma_{0:t_f}$ as:
\begin{align}
    \set{F}_\mu &\define \qty{\mu_0, \dots, \mu_{t_f} \in \set{R}^n\ \middle|\  (\mu_0, \mu_{t_f}) = (\mu_{\mathrm{ic}}, \mu_{\mathrm{tc}}), \text{\cref{eq:mean dynamics}}},\nonumber\\
    \set{F}_\Sigma &\define \qty{\Sigma_0, \dots, \Sigma_{t_f} \in \set{S}_+^n\ \middle| (\Sigma_0, \Sigma_{t_f}) = (\Sigma_{\mathrm{ic}}, \Sigma_{\mathrm{tc}}), \text{\cref{eq:cov dynamics}}}.\nonumber
\end{align}

\textbf{Cost Function.} Similarly, the cost function is approximated to second-order. Up to a constant, it is given by:
\begin{align}
\bar{c}_t(x, u) \define \frac{1}{2}(\transp{x} Q_t x + \transp{u} R u) + \transp{q}_t x.
\end{align}
where $Q_t \define \partial_x^2 c_t(\bar{x})$, and $q_t \define -Q_t \bar{x}_t$. The expected cost also factors into a mean component and steering component. Leading to two separate objectives:
\begin{align}
J_{\mu}(\mu_{0:t_f}, v_{0:t_f}) &\define \sum_{t = 0}^{t_f - 1} \bar{c}_{t}(\mu_t, v_t),\\
J_{\Sigma}(\Sigma_{0:t_f}, K_{0:t_f}) &\define \sum_{t = 0}^{t_f - 1} \trace(Q_t \Sigma_t) + \trace(R_t K_t \Sigma_t \transp{K}_t).\nonumber
\end{align}

\textbf{Chance Constraint.} The chance constraint \cref{eq:chance constraint} is, in general, intractable since it requires evaluating a multidimensional integral. We follow a standard approach based on Boole's inequality to produce a tractable conservative approximation. Specifically, for a state $X_t$, it holds that:
\begin{align}
\func{P}\qty{X_t \in \set{H}} &\leq \sum_{i = 1}^N \func{P}\qty{h_i(X_t) \leq 0}.
\end{align}
Therefore, the following constraint satisfies \cref{eq:chance constraint},
\begin{align}
\func{P}\qty{h_i(X_t) \leq 0} \leq \delta/N, && i = 1, \dots, N.
\end{align}
% Bounding the violation of all chance constraints is more conservative than the original chance constraint, but leads to a tractable single-dimensional integral.\footnote{Risk allocation methods can produce less conservative constraints \cite{Pilipovsky21}.}

Next, since $h_i(x)$ is, in general, nonlinear, an approximation is required to formulate a tractable constraint. Specifically, each nonlinear constraint is formulated as a half-plane constraint given by the linearization of $h_i(x)$, i.e.,
\begin{align}
\bar{h}_{i, t}(x) \define \transp{a}_{i, t} x + b_{i, t} \leq 0,
\end{align}
where $a_{i, t} \define \partial_x h_i(\bar{x}_{t}), b_{i, t} \define h(\bar{x}_t) - \transp{a}_{i, t}\bar{x}_t$
Since $X_t$ is Gaussian, each hyperplane constraint can be formulated as a deterministic function of $(\mu_t, \Sigma_t)$,
\begin{align}
\func{P}\qty{\transp{a}_{i, t} X_t + b_{i, t} \leq 0} \leq \delta' \iff g_{i, t}(\mu_t, \Sigma_t) \leq 0,
\end{align}
where:
\begin{align}
g_{i, t}(\mu_t, \Sigma_t) \define \transp{a}_{i, t} \mu_t + b_{i, t} + \inv{\Phi}\qty(1 - \delta')\sqrt{\transp{a}_{i, t}\Sigma_t a_{i, t}} \leq 0.\nonumber
\end{align}

Here, $\inv{\Phi}(\cdot)$ is the inverse of the Gaussian cumulative density function (CDF) and $\delta' \define \nicefrac{\delta}{N}$. In the desired case, $\delta' < 0.5$, the function $g_i(\mu, \Sigma)$ is concave in $\Sigma$, so the constraint is unfortunately still nonconvex. However, since it is a concave function, a linearization of $g_i(\mu, \Sigma)$ will provide a conservative approximation. Specifically, $g_i(\mu, \Sigma)$ is approximated by a first-order Taylor expansion,
\begin{align}
\bar{g}_{i, t}(\mu, \Sigma) \define c_{i, t} + \transp{w}_{i, t} \mu + \trace\qty(G_{i, t}\transp{\Sigma}),\nonumber
\end{align}
where $w_{i, t} \define \partial_\mu g_{i, t}(\bar{x}_t, \bar{\Sigma}_t), G_{i, t} \define \partial_\Sigma g_{i, t}(\bar{x}_t, \bar{\Sigma}_t),$ and $c_{i, t} = g_{i, t}(\bar{\mu}_t, \bar{\Sigma}_t) - \transp{w}_{i, t} \bar{x} - \trace(\smash{G_{i, t} \transp{\bar{\Sigma}}})$. Denote the set of means and covariance matrices satisfying the approximated constraint by: $\set{G}_{t} \define \{(\mu, \Sigma)\ |\  \bar{g}_{t}(\mu, \Sigma) \leq 0,\ i = 1, \dots, N \}$.

% \footnote{If trajectories are parameterized using $\Sigma_t^{\nicefrac{1}{2}}$, then the constraint becomes a convex second-order cone constraint. However, this is not possible using the chosen controller parameterization. See \cref{ssec:controller parameterization} for discussion.}

\textbf{Proximal Regularization.} Since the covariance steering problem being formed uses local approximations to the nonlinear dynamics and costs, it is necessary to control how far the solution deviates from the trajectory used by linearization. In this paper, a proximal method is used. Specifically, we introduce a quadratic proximal regularizer,
\begin{align}
J_{\mathrm{px}}(\mu_{0:t_f}, \Sigma_{0:t_f}) \define \sum_{t = 0}^{t_f} \frac{\alpha_\mu}{2} \norm{\mu_t - \bar{\mu}_t}_2^2 + \frac{\alpha_\Sigma}{2}\norm{\Sigma_t - \bar{\Sigma}_t}_{\mathrm{F}}^2,\nonumber
\end{align}
where $\norm{\cdot}_{\mathrm{F}}$ is the Frobenius norm. The parameters $\alpha_\mu, \alpha_\Sigma$ control the strength of the regularization.

\textbf{Local Covariance Steering Problem.} Following the preceding developments, we can formulate a local, convex approximation to \cref{eq:nonlinear cov steer prob} given a trajectory $(\bar{x}_{0:t_f}, \bar{u}_{0:t_f}, \bar{\Sigma}_{0:t_f})$:
\begin{mini}|s|
    {}
    {J_{\mu} + J_{\Sigma} + J_{\mathrm{px}},}
    {\label{eq:lq cov steer prob}}{}
    \addConstraint{(\mu_t, \Sigma_t) \in \set{G}_t,}{\ \mu_{0:t_f} \in \set{F}_\mu,}{\ \Sigma_{0:t_f} \in \set{F}_\Sigma}.
    %\addConstraint{}{}{}
\end{mini}

% {\color{red}
% \section{Distributed Optimization}
% \subsection{Alternating Direction Method of Multipliers}

% \subsection{Bregman Alternating Direction Method of Multipliers}

% }

% \newpage

% \newpage

\section{Operator Splitting Covariance Steering}

In this section, we explain how the ADMM is used to formulate an operator splitting scheme that solves the local covariance steering problem \cref{eq:lq cov steer prob}. In the absence of chance constraints, problem \cref{eq:lq cov steer prob} can be decoupled into a mean subproblem using the open-loop controls $v_{0:t_f-1}$ and a covariance one which uses the feedback matrices $K_{0:t_f-1}$. 
These two problems can then be solved in parallel. 
Nevertheless, even a single state chance constraint produces a coupling resulting in a sizeable SDP that cannot be parallelized \cite{Balci24}. In addition, using such an approach within an iterative linearization scheme would require feasible initializations and intermediate solutions during all iterations, which might significantly reduce the exploration capabilities of the algorithm.

To address these issues, we propose an operator splitting scheme using the ADMM to decouple chance constraint satisfaction from the two steering problems. This reformulation is performed by first introducing copies of the mean and covariance decision variables. The copy $(\mu^{\mathrm{s}}_{0:t_f}, \Sigma^{\mathrm{s}}_{0:t_f})$ is used for the steering problems, the copy $(\mu^{\mathrm{cc}}_{0:t_f}, \Sigma^{\mathrm{cc}}_{0:t_f})$ is used to enforce the chance constraints, and the copy $(\mu^{\mathrm{cn}}_{0:t_f}, \Sigma^{\mathrm{cn}}_{0:t_f})$ is used to ensure consensus. For notational convenience, we also define the following variable groups:
\begin{align}
z_\mu \define (v_{0:t_f}, \mu^{\mathrm{s}}_{0:t_f}), && z_\Sigma \define (K_{0:t_f}, \Sigma^{\mathrm{s}}_{0:t_f}),\\
z_{\mathrm{cc}} \define (\mu^{\mathrm{cc}}_{0:t_f}, \Sigma^{\mathrm{cc}}_{0:t_f}), && z_{\mathrm{cn}} \define (\mu^{\mathrm{cn}}_{0:t_f}, \Sigma^{\mathrm{cn}}_{0:t_f}).
\end{align}
Next, we form three objectives that include the constraints via indicator functions:
\begin{align}
\tilde{J}_{\mu}(z_\mu) &\define J_{\mu}(\mu_{0:t_f}, v_{0:t_f}) + I_{\set{F}_\mu}(\mu_{0:t_f}),\\
\tilde{J}_{\Sigma}(z_\Sigma) &\define J_{\Sigma}(\Sigma_{0:t_f}, K_{0:t_f}) + I_{\set{F}_\Sigma}(\Sigma_{0:t_f}),\nonumber\\
\tilde{J}_{\mathrm{cc}}(z_{\mathrm{cc}}) &\define I_{\set{G}_1}(\mu^{\mathrm{cc}}_{1}, \Sigma^{\mathrm{cc}}_{1}) + \dots + I_{\set{G}_{t_f}}(\mu^{\mathrm{cc}}_{t_f - 1}, \Sigma^{\mathrm{cc}}_{t_f - 1}).\nonumber
\end{align}
Then, we can reformulate problem \cref{eq:lq cov steer prob} as
\begin{mini}|s|
    {}{\tilde{J}_{\mu}(z_\mu) + \tilde{J}_{\Sigma}(z_\Sigma) + \tilde{J}_{\mathrm{cc}}(z_{\mathrm{cc}}) + J_{\mathrm{px}}(z_{\mathrm{cn}}),}
    {\label{eq:admm cov steer prob}}{}
    \addConstraint{\mu^{\mathrm{s}}_{0:t_f} = \mu^{\mathrm{cn}}_{0:t_f},}{\quad \Sigma^{\mathrm{s}}_{0:t_f} = \Sigma^{\mathrm{cn}}_{0:t_f}}{}
    \addConstraint{\mu^{\mathrm{cc}}_{0:t_f} = \mu^{\mathrm{cn}}_{0:t_f},}{\quad \Sigma^{\mathrm{cc}}_{0:t_f} = \Sigma^{\mathrm{cn}}_{0:t_f}}{}.
\end{mini}
Note that since the new problem consists of separable objective functions and linear equality constraints, we can derive an ADMM scheme for solving it. The (scaled) augmented Largrangian (AL) of \cref{eq:admm cov steer prob} is given by
\begin{align}
\mathcal{L}_\rho & = \tilde{J}_{\mu}(z_\mu) + \tilde{J}_{\Sigma}(z_\Sigma) + \tilde{J}_{\mathrm{cc}}(z_{\mathrm{cc}}) + J_{\mathrm{px}}(z_{\mathrm{cn}})
\nonumber
\\
& + \frac{\rho}{2} \Big[ \left\| \mu^{\mathrm{s}} - \mu^{\mathrm{cn}} + \lambda_1 \right\|_2^2
+ \left\| \mu^{\mathrm{cc}} - \mu^{\mathrm{cn}} + \lambda_2 \right\|_2^2
\\
& + \left\| \Sigma^{\mathrm{s}} - \Sigma^{\mathrm{cn}} + \Lambda_1 \right\|_\mathrm{F}^2
+ \left\| \Sigma^{\mathrm{cc}} - \Sigma^{\mathrm{cn}} + \Lambda_2 \right\|_\mathrm{F}^2 \Big],
\nonumber
\end{align}
where $\rho$ is a consensus penalty parameter and $\lambda_1, \lambda_2, \Lambda_1, \Lambda_2$ are the dual variables for the corresponding equality constraints. In the previous expression, the time indices have been temporarily dropped for notational brevity.
% is in the form of \cref{eq:operator splitting} and can be solved via the ADMM.
Then, we can derive the following two-block ADMM algorithm, where the first block consists of minimizing $\mathcal{L}$ w.r.t. $z_\mu$, $z_\Sigma$, $z_{\mathrm{cc}}$, leading to the subproblems
\begin{align}
& z_\mu \leftarrow {\operatorname{prox}}_{\rho, \tilde{J}_\mu} [\mu^{\mathrm{cn}} - \lambda_1],
\quad 
z_\Sigma \leftarrow {\operatorname{prox}}_{\rho, \tilde{J}_\Sigma} [\Sigma^{\mathrm{cn}} - \Lambda_1],
\nonumber
\\
& z_\mathrm{cc} \leftarrow {\operatorname{prox}}_{\rho, \tilde{J}_\mathrm{cc}} [(\mu^{\mathrm{cn}}, \Sigma^{\mathrm{cn}}) - (\lambda_2, \Lambda_2)],
\label{eq: admm block 1}
\end{align}
which can all be executed in parallel. In addition, note that the subproblem solving for $z_{\mathrm{cc}}$ can be further parallelized for all time instants $t = t_0, \dots, t_f$. The second ADMM block involves minimizing the AL w.r.t. $z_\mathrm{cn}$, which leads to the following averaging steps
\begin{equation}
\mu^{\mathrm{cn}} \leftarrow (\mu^{\mathrm{s}} + \mu^{\mathrm{cc}})/2, 
\quad 
\Sigma^{\mathrm{cn}} \leftarrow (\Sigma^{\mathrm{s}} + \Sigma^{\mathrm{cc}})/2.
\label{eq: admm block 2}
\end{equation}
Finally, the dual variables are updated as follows
\begin{align}
\lambda_1 & \leftarrow \lambda_1 + \rho (\mu^{\mathrm{s}} - \mu^{\mathrm{cn}}), 
~ 
~ \lambda_2 \leftarrow \lambda_2 + \rho (\mu^{\mathrm{cc}} - \mu^{\mathrm{cn}}), 
\label{eq: admm block dual}
\\
\Lambda_1 & \leftarrow \Lambda_1 + \rho (\Sigma^{\mathrm{s}} - \Sigma^{\mathrm{cn}}), 
~ 
\Lambda_2 \leftarrow \Lambda_2 + \rho (\Sigma^{\mathrm{cc}} - \Sigma^{\mathrm{cn}}). 
\nonumber
\end{align}
%
% Importantly, when solving via the ADMM, the mean steering, covariance steering, and projection onto the chance constraint at each time step \emph{can all be executed in parallel}. This is critical to the computational performance improvements of the operator splitting method. 
Note that the consensus constraints in \cref{eq:admm cov steer prob} relax the chance and dynamics constraints. Thus, iterates of the resulting ADMM algorithm are allowed to be temporarily infeasible. In practice, this allows the proposed algorithm to find feasible solutions to \cref{eq:lq cov steer prob} when common SDP solvers might fail.

\subsection{Updating the Nominal Trajectory}

\label{ssec:trajectory update}

% {\color{red} Show more simulations. These simulations should include:

% \begin{itemize} 
% \item 
% The same systems (car and drone) but more environments and tasks.
% \item Comparisons with Bregman ADMM. These comparisons should include performance (initial objective function) and speed of convergence. 
% \end{itemize}
% } 

A procedure (i.e., \texttt{ForwardPass} in \cref{alg:csos})  to update $\tau$ once the ADMM converges is needed to complete the algorithm. This update can be achieved using various methods to approximately compute the mean and covariance corresponding to the nonlinear dynamics \cref{eq:dynamics}. One possible option available is to estimate these quantities from many samples of \cref{eq:dynamics} using the most recently computed controller parameters. While sampling can be performed in parallel, larger state spaces and longer time horizons can still require a prohibitive number of samples. In such cases, if the process noise is small and the regularizer is well-tuned, $\bar{x}_{0:t_f}, \bar{u}_{0:t_f}$ can be estimated by simply propagating the mean dynamics (i.e., \cref{eq:dynamics} with the process noise term removed) using the new control parameters and $\Sigma_{0:t_f}^\mathrm{s}$ as an approximation for $\bar{\Sigma}_{0:t_f}$. The latter method was found to work quite well and is the method used in \cref{sec:simulations} and \cref{sec:hardware}.

\subsection{Convergence and Stopping Criteria}

Our algorithm is a two-level optimization scheme, where the outer level corresponds to a trust-region successive convexification approach, while the inner level involves solving convex SDPs. The convergence of the inner problems \eqref{eq:admm cov steer prob} to their optimal solution is straightforward to establish using established results about the convergence of ADMM due to the convexity of the cost terms and the full-rank structure of the linear equality (identity) constraint matrices \cite{Deng16}. Given the guaranteed convergence of the inner convex problems, we can then employ a convergence analysis for trust-region sequential convexification methods (e.g. \cite{Mao16}) to prove the convergence of the outer level, and thus the overall algorithm to a stationary point of the original non-convex optimization problem \eqref{eq:nonlinear cov steer prob}. A formal exposition of this convergence analysis will be illustrated in an extended version of this manuscript.

Since our algorithm allows for temporary infeasibility during intermediate steps, it is important to properly establish a stopping criteria. In the case of the inner-level problem, the primal residual of the ADMM problem can be used as the stopping criterion. The problem is considered to have fully converged when the residuals are below a predetermined tolerance. In a similar manner, for the outer-level problem, we can use the trust region residuals as a stopping criteria.  In practice, we  observed that a fixed number of inner-level iterations and using constraint satisfaction as a criterion to terminate the outer problem iterations yields safe controls. However, to satisfy the convergence guarantees, the inner and outer problems must both run until they fully converge.

\begin{algorithm}[b]
    \caption{Operator Splitting Scheme}
    \label{alg:os}
    \begin{algorithmic}[1] % The number tells where the line numbering should start
    \Require Initial $z_\mu, z_\Sigma, z_{\mathrm{cc}}, z_{\mathrm{cn}}$, $\lambda_1, \lambda_2, \Lambda_1, \Lambda_2$, $\rho$.
    \Repeat
    \State $z_\mu, z_\Sigma, z_{\mathrm{cc}} \gets$ Solve \cref{eq: admm block 1}. \quad \quad \quad \texttt{\# In parallel}
    \State $z_{\mathrm{cn}} \gets$ Update with \cref{eq: admm block 2}. \quad \quad \quad \texttt{\# In parallel}
    \State $\lambda_1, \lambda_2, \Lambda_1, \Lambda_2 \gets$ Update with \cref{eq: admm block dual}.
    \Until{Converged}
    \end{algorithmic}
\end{algorithm}

\begin{algorithm}[b]
    \caption{Covariance Steering via Operator Splitting}
    \label{alg:csos}
    \begin{algorithmic}[1] % The number tells where the line numbering should start
    \Require Initial trajectory $\tau^{(0)} \define (\bar{x}^{(0)}_{0:t_f}, \bar{u}_{0:t_f}^{(0)}, \bar{\Sigma}_{0:t_f}^{(0)})$.
    \Repeat
    \State $A_{0:t_f}^{(k)}, B_{0:t_f}^{(k)}, d_{0:t_f}^{(k)}, Q_{0:t_f}^{(k)}, q_{0:t_f}^{(k)}, c_{i, 0:t_f}^{(k)}, w_{i, 0:t_f}^{(k)}, G_{i, 0:t_f}^{(k)}$ 
    
    $\gets$ \texttt{TaylorExpansions}$(\tau^{(k)})$.
    \State $\mu_{0:t_f}^{(k)}, \Sigma_{0:t_f}^{(k)}, v_{0:t_f}^{(k)}, K_{0:t_f}^{(k)} \gets$ Solve with Alg. \ref{alg:os}.
%
    % \State $(x^{[j]}_{0:t_f}, u_{0:t_f - 1}^{[j]})_{j = 1}^M \gets$ Sample $\dist{N}(\mu_{\mathrm{ic}}, \Sigma_{\mathrm{ic}})$ and \cref{eq:dynamics}.
%
    \State $\tau^{(k + 1)} \gets$ \texttt{ForwardPass}$(\mu_{0:t_f}^{(k)}, \Sigma_{0:t_f}^{(k)}, v_{0:t_f}^{(k)}, K_{0:t_f}^{(k)})$
    \Until{Converged}
    \end{algorithmic}
% \vspace{-0.5cm}
\end{algorithm}

\section{Simulation Experiments}
\label{sec:simulations}
% Then  To show the performance of our method we have performed numerical experiments on two non-linear systems - unicycle and quadrotor. 

\begin{figure*}
\captionsetup[subfigure]{labelformat=empty}
\centering
\subfloat[]{\centering
    \includegraphics[width=0.65\columnwidth, trim={0cm 0cm 0cm 0cm},clip]{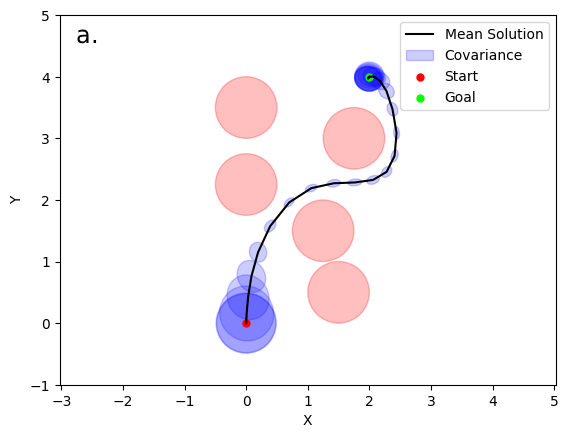}
    %\caption{Solution to Chance constraint CSP for a double integrator system using baseline SDP method}
    \label{fig:linear_sdp}
}
%no space
\hfill
\subfloat[]{\centering
    \includegraphics[width=0.65\columnwidth, trim={0cm 0cm 0cm 0cm},clip]{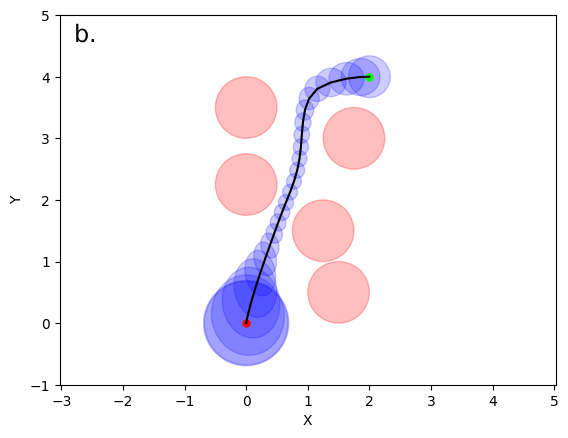}
    %\caption{Solution to Chance constraint CSP for a double integrator system using our operator splitting based method}
    \label{fig:linear_admm}
}
\hfill
\subfloat[]{
\includegraphics[width=0.635\columnwidth, trim={0cm 0cm 0cm 0cm},clip]{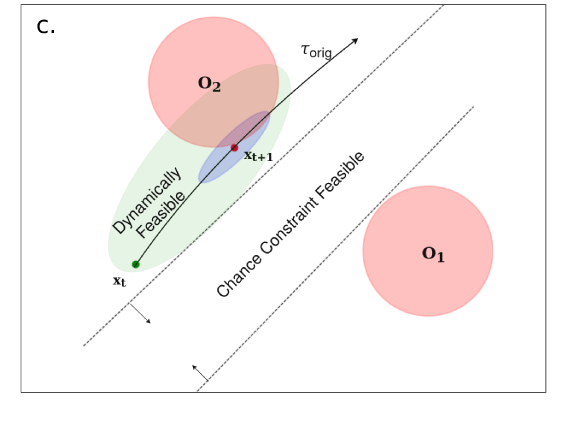}
%    \caption{Example of infeasible SDP problem. }
\label{fig: issue}
}
\vspace{-0.6cm}
\caption{The results of the ablation study conducted in \cref{ssec:ablation}. Our algorithm found a solution that exhibits a lower cost while also satisfying tighter safety constraints.  a) The solution found by the baseline method. Notice that, despite a loser safety requirement, the covariances are much tighter and the trajectory takes larger steps compared to our solution. This is an artifact of the optimizer stuck in a bad local minima where the only feasible solution requires precisely placed means and covariances. b) The solution found by our algorithm. c) This schematic visually depicts why traditional solvers struggle to solve \cref{eq:lq cov steer prob} as part of iterative algorithms. The trajectory $\tau_{\mathrm{orig}}$ is the trajectory around which the approximations occur. The half-planes shown correspond to the linearized chance constraints. The set of points that are feasible with respect to the chance constraint and those that are dynamically feasible cannot both be satisfied.}
\label{fig:csetup}
\end{figure*}

This section demonstrates the effectiveness of our method on controlling robotic systems. We start with illustrating a task of safely navigating a unicycle while avoiding multiple obstacles (\cref{sec: sim results - unicycle}). Then, a quadrotor task that features significantly more challenging dynamics along with an obstacle is addressed (\cref{sec: sim results - quadrotor}). Finally, an ablation study is performed in \cref{ssec:ablation} that compares the operator splitting method with solving \cref{eq:lq cov steer prob} as a single SDP. 
% Additionally, to show the capabilities of our algorithm, we have also tested our method on the Robotarium \cite{robotarium} hardware platform. 
In our experiments, to enforce obstacle avoidance we use a signed distance function to each obstacle as the constraint $h_i(X_t)$. 
All optimization problems are solved with CVXPY \cite{Diamond16} and MOSEK on Intel i9-14900K. \cite{mosek_python}.

\subsection{Unicycle}
\label{sec: sim results - unicycle}

\begin{figure*}[b]{%
    \includegraphics[width=0.2\textwidth]{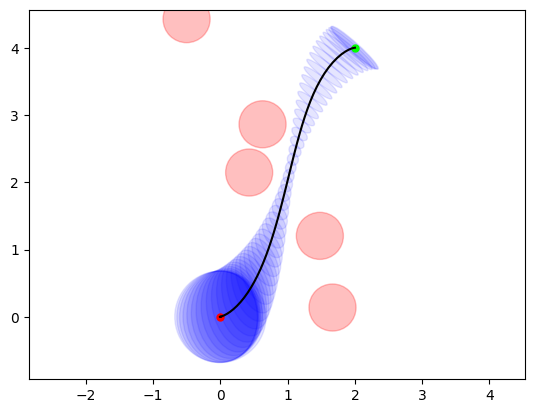}%
    \includegraphics[width=0.2\textwidth]{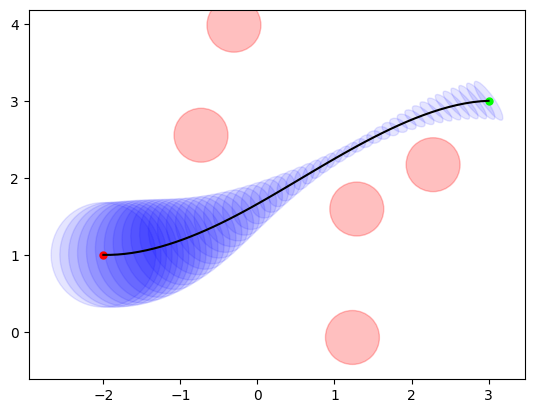}%
    \includegraphics[width=0.2\textwidth]{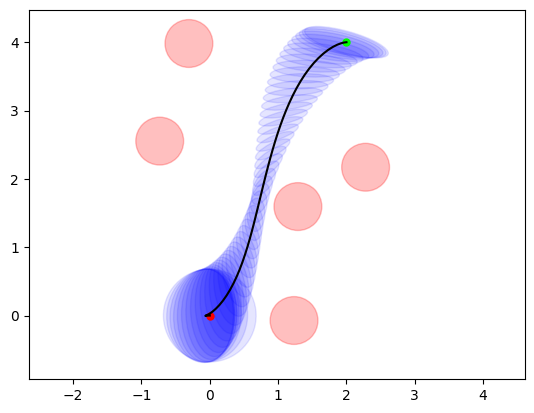}%
    \includegraphics[width=0.2\textwidth]{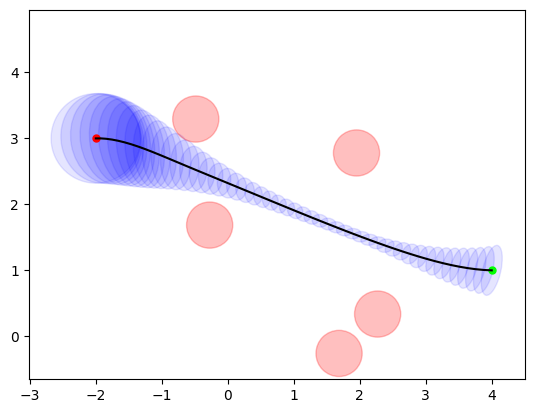}%
    \includegraphics[width=0.2\textwidth]{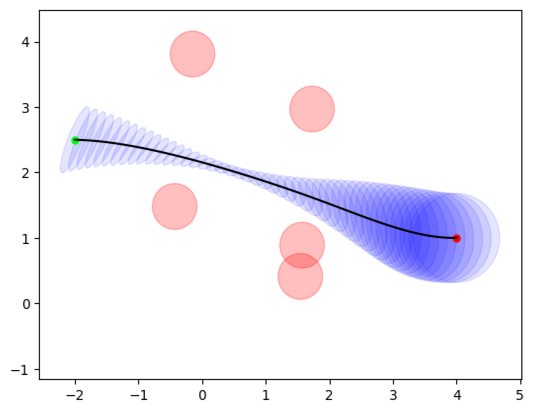}%
    \label{fig: random unicyle}

    \vspace{-0.4cm}
    \caption{Five example environments used in the unicycle experiment described in \cref{sec: sim results - unicycle}. Each environment was formed by sampling perturbations for the obstacles from the configuration in \cref{fig:unicycle}. In each case, our algorithm easily found solutions that feature high probabilities of safety.}
}
\end{figure*}

We first consider a unicycle robot with state $x = (\mathrm{x}, \mathrm{y}, \theta)$ and control input $u = (v, \omega)$. The time step is $\Delta t = 0.1$ and $t_f = 50$. The initial and target covariances are specified through $\Sigma_0=0.1I_3$ and $\Sigma_{t_f} \preceq 0.1I_3$, while $D_t=0.01I_3$. We have chosen a quadratic cost function with $Q=0.001I_3$ and $R=0.1I_2$. 
For this experiment, we placed 5 obstacles as shown in \cref{fig:unicycle}. Optimal planning in this environment requires navigating through a somewhat narrow central channel and appropriately reducing the covariance. The covariance is then free to expand to satisfy the terminal constraint when in the open space near the terminal condition. Then, to demonstrate the effectiveness of our algorithm, we also tested it in 10 different environment configurations. We warm-started each environment with an initial solution to aid in faster convergence. While our method can handle poor initializations, the outer optimization process may be slow to converge. In practical applications, a better method to initialize the system is to use a motion planning algorithm, like RRT$^\star$, to find a feasible initial trajectory.

The statistics of these experiments averaged over all 10 environments is shown in \cref{tabl:results}. The solution for each environment was obtained by running \cref{alg:csos} for 10 iterations and solving \cref{eq:lq cov steer prob} using 15 ADMM iterations. On average, our method required 2.67 seconds for computation. Furthermore, for each of the solutions, we empirically compute the approximate objective value from the last solution of \cref{eq:lq cov steer prob} and the probability of safety from 200 sample trajectories each. Example environments along with the final mean and covariance found by our algorithm are shown in \cref{fig: random unicyle}. Sampled solution paths are depicted in \cref{fig:unicycle}.
 
% {\color{red}The solution was obtained by running Algorithm \ref{alg:csos} for 10 iterations and solving \cref{eq:lq cov steer prob} using 15 iterations of ADMM. For this task, our method required 2.13 seconds for computations. Table \ref{tabl:results} compares the approximate objective value from the last solution of \cref{eq:lq cov steer prob} with the mean cost from 200 sample trajectories. The solution and sample trajectories of the unicycle are depicted in \cref{fig:unicycle}.}

% \begin{figure}[t]
%     \centering
%     \includegraphics[width=\columnwidth]{images/unicyle.png}
%     \caption{Simulation results of unicycle using our algorithm. The results are from 200 Monte-Carlo run, five of which are shown in the figure.}
%     \label{fig:unicycle}
% \end{figure}

\begin{table}[]
    \centering
\begin{tabular}{@{}lcccc@{}}
\toprule
Experiment &           & Cost   & Safety Prob.  \\ \midrule
Unicycle     & Optimizer & 7.93 & 0.990 &   \\
             & Estimated & 8.29 & 0.975 &  \\ \midrule[0.05em]
Quadrotor  & Optimizer & 762.00 & 0.990 &  \\
           & Estimated & 821.38 & 1.000 &  \\ \midrule[0.05em]
Robotorium & Optimizer & 7.94 & 0.900 &  \\
           & Estimated & 8.06 & 0.970 &  \\ \bottomrule
\end{tabular}
\caption{Summary of Experimental Results}
\label{tabl:results}
\vspace{-0.9cm}
\end{table}

\subsection{Quadrotor}
\label{sec: sim results - quadrotor}
\begin{figure}[t]
\vspace{-0.6cm}
    \centering
    \includegraphics[width=0.65\columnwidth]{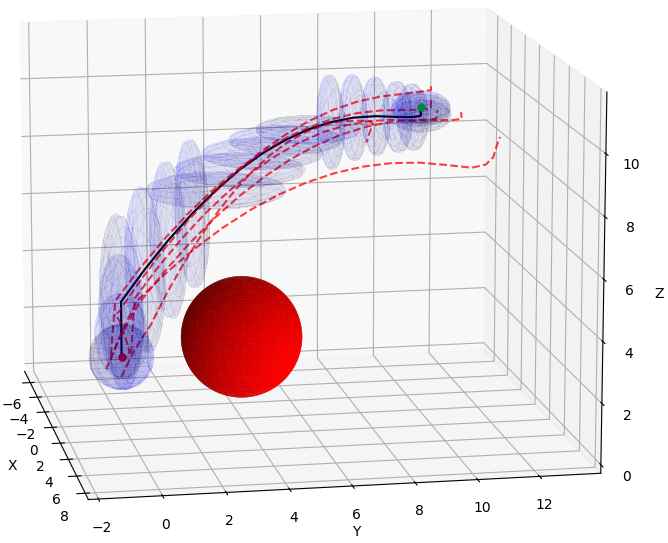}
    \vspace{-0.3cm}
    \caption{Simulation results of quadrotor navigating around a spherical obstacle. Five trajectory samples are plotted as red dashed lines.}
    \label{fig:quadrotor}
\end{figure}

A similar experiment is performed for navigating a 3D quadrotor (12 states, 4 control inputs) around a spherical obstacle. Here, $\Delta t=0.1$, $t_f=20$, $\Sigma_0=0.1I_{12}$, $\Sigma_{t_f} \preceq 0.1I_{12}$ and control noise $D_t = 0.1I_4$. Estimated costs from the value of the last approximation and sample trajectories are available in \cref{tabl:results}. \cref{alg:csos} was run for 10 iterations, while the number of ADMM iterations is also set to 10. The solution was computed in 25.68 seconds. To avoid numerical issues resulting from large angular values,
% \footnote{This issue stems from using an Euler angle parameterization for the rotational variables.}
the initial state is sampled from a truncated Gaussian. \cref{fig:quadrotor} shows the resulting mean trajectory and a few examples of simulated trajectories. Our method successfully satisfies the required probability of safety while reaching the goal location. 

% \begin{figure}[t]
%     \centering
%     \includegraphics[width=0.8\columnwidth]{images/infeasibility_sdp.png}
%     \caption{This schematic demonstrates why traditional solvers struggle to to solve \cref{eq:lq cov steer prob} as part of iterative algorithms. The trajectory $\tau_{orig}$ refers to the trajectory around about which the approximations occur. The half-planes shown correspond to the linearized chance constraints. The set of points that are feasible with respect to the chance constraint and those that are dynamically feasible cannot be simultaneously satisfied. This failure mode can occur even in the simple case of a double integrator when the trajectory is strictly feasible.}
%     \label{fig:infeasibility_sdp}
% \end{figure}

\subsection{Comparison with Baseline Algorithm}
\label{ssec:ablation}

To analyze the performance capabilities of our method, it is important to compare our method with a baseline where \cref{eq:lq cov steer prob} is treated as a single SDP without operator splitting. To simplify the comparison, we use a double integrator with $\Delta t = 0.2$, $\Sigma_0 = 0.1I_4$, $\Sigma_{t_f}=0.05I_4$ and $D_t = B_t$. The cost function is quadratic with $Q=0.01I_4$ and $R=0.005I_2$. 
Both methods were run for 10 iterations with warm starting. In our algorithm, ADMM was run for 15 iterations. Table \ref{tabl:sdp_admm_results} summarizes the results from our experiment. The baseline tends to converge more quickly but fails to generate a feasible solution when the required probability of safety is $\delta' = 0.99$ even if the dynamics are linear. In particular, it is only able to find a feasible solution when $\delta' = 0.9$. On the contrary, our method found a solution with the tighter constraint $\delta' = 0.99$ that also features a lower cost. Figures \ref{fig:linear_sdp} and \ref{fig:linear_admm} show the resulting trajectories from the two methods.

Since our method allows the iterates to temporarily relax the constraints, it can find feasible trajectories in cases where the baseline method fails since the latter always requires all constraints to be satisfied simultaneously. This poses a problem when constructing the convex approximation \cref{eq:lq cov steer prob} about a new trajectory. Essentially, when linearizing the obstacle constraints about a new trajectory, the intersection of the set of dynamically feasible points and points that satisfy the chance constraints becomes disjoint. \cref{fig: issue} highlights an occurrence of this limitation. On the other hand, our method circumvents this issue since the dynamical and chance constraints are not required to be simultaneously satisfied prior to convergence.

% \begin{figure}[t]
%     \centering
%     \includegraphics[width=\columnwidth]{images/linear_sdp2.png}
%     \caption{Solution to Chance constraint CSP for a double integrator system using baseline SDP method}
%     \label{fig:linear_sdp}
% \end{figure}

% \begin{figure}[t]
%     \centering
%     \includegraphics[width=\columnwidth]{images/linear_admm2.png}
%     \caption{Solution to Chance constraint CSP for a double integrator system using our operator splitting based method}
%     \label{fig:linear_admm}
% \end{figure}

% \begin{table}[]
%     \centering
%     \begin{tabular}{|c|c|c|c|}
%         \hline
%           $N=25$& Cost & Solve Time(sec) & Max Safe Prob.  \\
%           \hline
%           Baseline &  2.9436 & \textbf{0.2972} & 0.90 \\
%           \hline
%           Our Method & \textbf{2.5977} & 2.6451 & \textbf{0.99} \\
%           \hline
%     \end{tabular}
%     \caption{Baseline vs ADMM}
%     \label{tab:sdp_vs_admm}
% \end{table}

\begin{table}[]
\vspace{0.25cm}
    \centering
    \begin{tabular}{@{}lccc@{}}
    \toprule
               & Solve Time & Cost   & Safety Prob.  \\ \midrule 
    Baseline      & No Solution              & -  &  0.99\\
    Baseline      & \textbf{0.2972}              & 5.8860  &  0.90\\
    Our Algorithm & 2.6451                       & \textbf{5.1954} &  \textbf{0.99}\\
    \bottomrule
    \end{tabular}
    \caption{Comparison of Our Algorithm with Baseline}
    \label{tabl:sdp_admm_results}
\vspace{-1cm}
\end{table}

\section{Hardware Experiments}
\label{sec:hardware}
\begin{figure}[t]
    \vspace{0.25cm}
    \centering
    \includegraphics[width=0.9\columnwidth]{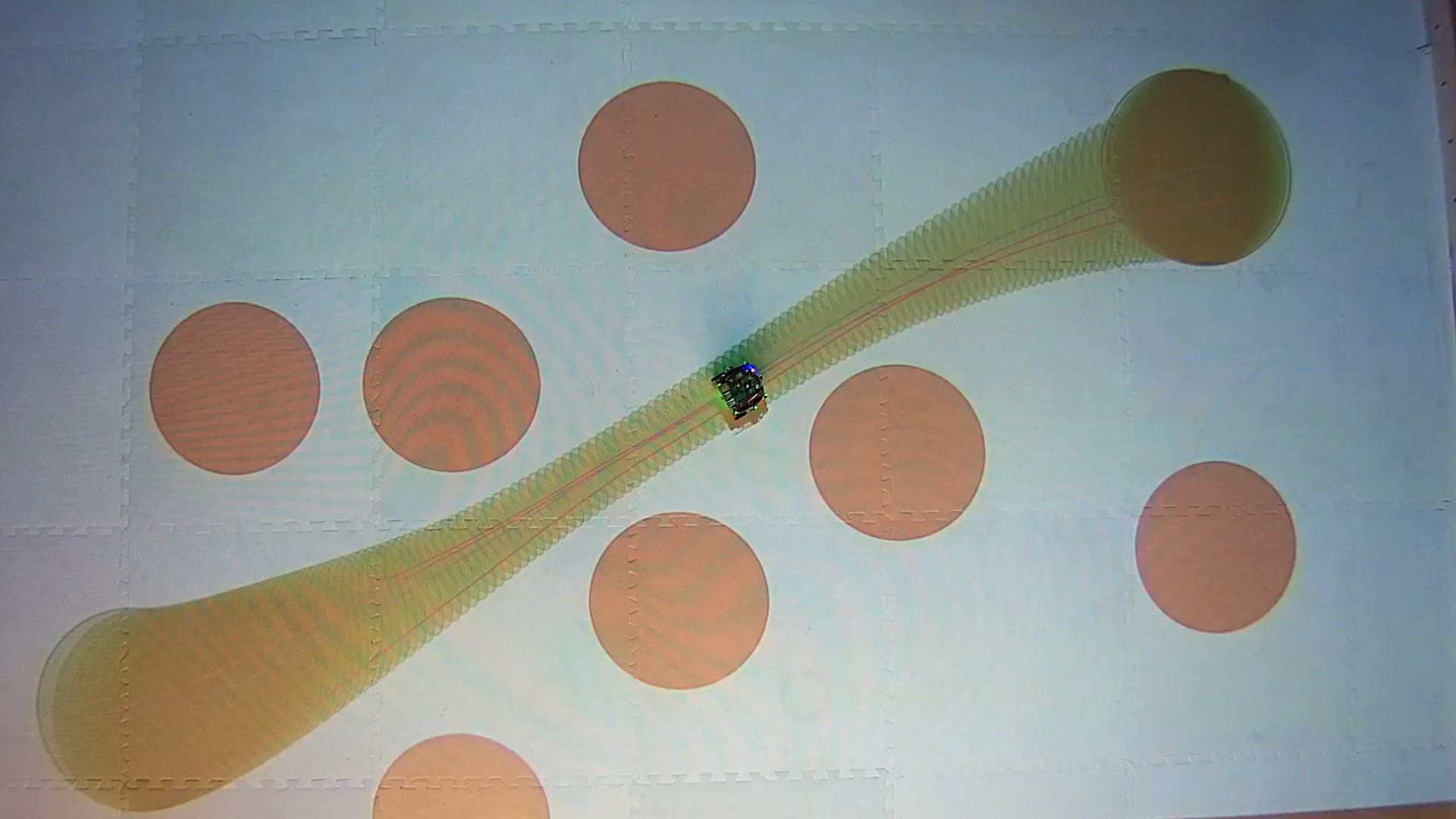}
    \caption{Example run of Robotarium GritsBot \cite{Pickem17} navigating an obstacle-rich environment. The feedback controller produced by our algorithm allowed the GritsBot to effectively navigate the environment. In 100 trials, only 3 violated the obstacle constraints.}
    \label{fig:robotarium}
    \vspace{-0.6cm}
\end{figure}

To show the usefulness of the controllers synthesized by our algorithm, we conduct a hardware experiment on the Robotarium platform \cite{Pickem17}, using a unicycle robot with the same unicycle dynamics as in Section \ref{sec: sim results - unicycle}. Figure \ref{fig:robotarium} shows the experimental setup, while the reader is encouraged to watch the supplementary video\footnote{Video: \url{https://youtu.be/UCyYcDITO2Q}} for a full demonstration of the experiments. We have set $\Delta t = 0.2$, $t_f = 200$, an initial covariance $\Sigma_0 = 0.05I_3$, terminal covariance $\Sigma_{t_f}=0.05I_3$ and a position noise $D_t = 0.01I_3$. For simplicity we have again chosen a quadratic cost with $Q=0.01I_3$ and $R=0.1I_2$. In addition to the obstacle constraints, we enforce a conservative wheel velocity constraints, $-b_{max} \leq G u_t \leq b_{max}$ while solving the mean steering problem in \cref{eq: admm block 1}, where $G=(R/2) [2,L; 2,L]$,  $b_{max}=[7;7]  \ \mathrm{rad}/\mathrm{s}^2$, wheel radius $R = 0.016\mathrm{m}$ and axle length $L=0.11 \mathrm{m}$. The solution was obtained by running 20 iterations of Algorithm \ref{alg:csos} and using 10 ADMM iterations to solve \cref{eq:lq cov steer prob}. Our algorithm took 19.03 seconds to run. The Robotarium platform was used to perform 100 trials with different initial conditions sampled and the results are shown in \cref{tabl:results}. As shown, the proposed controller exhibits reliable performance on a real robotic system. During the 100 runs, only 3 runs violated the safety constraint.

\section{Conclusion}
This article demonstrated that the proposed operator splitting method for solving nonlinear CSPs is effective at finding solutions with better performance and safety than extant methods. This improvement is due to the fact that the algorithm is tolerant to infeasible iterates during the optimization feature that often appear due to the sensitivity and conservatism of the necessary approximations. The advantages of our method were demonstrated on a variety of tasks including both simulation and hardware experiments.

There is a variety of possible extensions and applications for the proposed distribution steering methodology. One promising direction would be to replace the standard ADMM scheme with other variants that are intended for handling probability distributions such as Bregman ADMM \cite{Wang14}. Combining our algorithm with sampling-based motion planning \cite{Lavalle06} for finding safe plans, similar in spirit to LQR-Trees \cite{Tedrake10}, could also be an interesting area of research. Finally, we plan to explore integrating the proposed approach into learning-to-optimize frameworks for operator splitting methods \cite{Sambharya24, Saravanos25} to further enhance the computational efficiency and scalability of our algorithm.
% The latter would be beneficial since our algorithm is demonstrated to handle the nonlinear state constraints (i.e., obstacles) with which extant methods struggle.

\printbibliography
% \bibliographystyle{ieeetr}
% \bibliography{bibliography}

\end{document}